\documentclass[letterpaper, 10 pt, conference]{ieeeconf}  

\IEEEoverridecommandlockouts                              

\usepackage{times}
\usepackage{epsfig}
\usepackage{graphicx}
\usepackage{amsmath}
\usepackage{amssymb}
\usepackage{cite}
\usepackage{xcolor}
\usepackage{multirow}
\usepackage{booktabs}
\usepackage{array}
\usepackage{algorithm}
\usepackage[noend]{algpseudocode}
\usepackage{makecell}
\usepackage{paralist}
\graphicspath{{imgs/}}
\usepackage{soul}
\usepackage{caption}
\usepackage{pifont}
\usepackage{hyperref}
\usepackage{subcaption}
\usepackage[font=footnotesize]{caption} 
\makeatletter
\g@addto@macro\normalsize{%
  \setlength\abovedisplayskip{4pt}
  \setlength\belowdisplayskip{4pt}
  \setlength\abovedisplayshortskip{4pt}
  \setlength\belowdisplayshortskip{4pt}
}
\makeatother

\newcommand{\stxt}[1]{\textcolor{orange}{{}}}

\DeclareMathOperator{\Tr}{Tr}
\newcommand{\mymat}[1]{\mathbf{#1}}
\newcommand{\myset}[1]{\mathcal{#1}}
\newcommand{\mypcd}[1]{\textrm{\upshape #1}}
\newcommand{\myvec}[1]{\vec{#1}}

\title{\LARGE \bf
Category-Level Global Camera Pose Estimation with Multi-Hypothesis Point Cloud Correspondences
}

\author{Jun-Jee Chao$^{1}$, Selim Engin$^{1}$, Nicolai H{\"a}ni$^{1}$ and Volkan Isler$^{1}$
\thanks{$^{1}$ Department of Computer Science,
        University of Minnesota, USA
        \tt\small {\{chao0107, engin003, haeni001, isler\}@umn.edu}}%
}

\begin{document}

\maketitle
\thispagestyle{empty}
\pagestyle{empty}

\begin{abstract}
Correspondence search is an essential step in rigid point cloud registration algorithms. Most methods maintain a single correspondence at each step and gradually remove wrong correspondances. However, building one-to-one correspondence with hard assignments is extremely difficult, especially when matching two point clouds with many locally similar features. This paper proposes an optimization method that retains all possible correspondences for each keypoint when matching a partial point cloud to a complete point cloud. These uncertain correspondences are then gradually updated with the estimated rigid transformation by considering the matching cost. Moreover, we propose a new point feature descriptor that measures the similarity between local point cloud regions. Extensive experiments show that our method outperforms the state-of-the-art (SoTA) methods even when matching different objects within the same category. Notably, our method outperforms the SoTA methods when registering real-world noisy depth images to a template shape by up to $20\%$ performance. 
\end{abstract}

\section{Introduction}

Given an image of an object from a known category, computing the camera pose with respect to the object frame is a challenging problem in robot localization~\cite{zhang2019eye}, robotic manipulation~\cite{correll2016analysis} and object inspection~\cite{zollini2020uav}. This problem is even more challenging when we are given neither two views of the same scene nor the exact 3D model corresponding to the object. In our setting, the template shape can be significantly different from the observed object in terms of its geometry \emph{and} appearance. For example, it can be a simple synthetic model of a car that is significantly different from the observed real car.

To solve this novel pose estimation problem, new methods are needed:
Traditional Structure from Motion (SfM) explores visual features like SIFT~\cite{lowe2004distinctive} to establish correspondences between sparse views, but some overlap is required. Moreover, these features rely heavily on visual similarity. However, in our case, the observed object might have a completely different texture than the template shape from the synthetic domain. Point cloud registration methods such as the Iterative Closest Point (ICP) algorithm~\cite{besl1992method,segal2009generalized} provide an alternative way to estimate the camera pose by avoiding the domain gap introduced by texture. However, ICP is sensitive to initialization and can get trapped in local minima. Point feature descriptors are a way to mitigate the initialization and local minima problem by matching keypoints with distinctive local features~\cite{rusu2008aligning,rusu2009fast,zhou2016fast}.

Recently, learning-based methods have been proposed to address the limitations of these traditional methods. Instead of requiring view overlap for feature matching, deep learning approaches learn a direct mapping from images to poses~\cite{schwarz2015rgb, kendall2015posenet,kendall2017geometric}. More recently, deep learning approaches have been used to build 2D-3D correspondences~\cite{tremblay2018deep,tekin2018real,suwajanakorn2018discovery,pavlakos20176}. Nevertheless, these learning-based methods suffer from out-of-distribution samples during test time and are sensitive to texture differences. Besides learning from textures, deep learning is used to improve traditional point feature descriptors~\cite{yuan2020deepgmr, ginzburg2022deep,wang2019deep, yew2020-RPMNet, wu2021feature, lee2021deeppro, Fu2021RGM, yaoki2019pointnetlk}. However, learning-based methods require a large amount of labeled data which is not always applicable for real-world applications. Moreover, these methods usually require training on a specific category or dataset. In contrast, our proposed method requires only one point cloud from the target category, and is less sensitive to geometry difference and noise. We show that our method generalizes easily across different datasets and demonstrate its direct applicability to robotic grasping as a representative task. 


Our work follows the classical pipeline of building point correspondences with local features but adds a number of novel ideas. A fundamental limitation of existing optimization-based methods is that they build one-to-one correspondences between the observed partial point cloud and the complete point cloud. However, the point distribution between the complete and partial point clouds differs, making matching between local descriptors nontrivial. Moreover, our observed point cloud and the given template shape can have different geometry, which makes building one-to-one correspondences in a single shot more difficult. Our method addresses these issues as follows:

\begin{compactitem} 
\item  Instead of building one-to-one matching between keypoint features, we consider multiple hypotheses and design an optimization method to handle one-to-many soft feature assignments.
\item To account for different point distributions between the complete and partial point clouds, we introduce a partitioning scheme to register the observed partial depth scans to the provided template shape together with a novel keypoint descriptor that captures point distribution statistics in a local neighborhood. 
\item We show that our proposed approach can act as a drop-in replacement to existing feature matching methods and improves performance by up to 50\% on synthetic data.
\item We show that our method achieves comparable results with learning-based methods on synthetic data and outperforms them by up to 20\% on noisy real-world data. Moreover, without requiring any training, our method generalizes better across different datasets and objects.
\end{compactitem}

Experiments on synthetic data, real-world benchmarks, and robotic applications are conducted to justify our claims.

\section{Related Work}

This section reviews existing approaches for point cloud registration and category-level object pose estimation.

\textbf{Point cloud registration:}
The classical ICP~\cite{besl1992method,segal2009generalized} iteratively solves for the relative pose between two point clouds by alternating between closest point search and transformation estimation. 
Due to its myopic nature, it is well known that ICP can get stuck in local minima, especially when the two point-sets are not identical. 
Although newer variants have been proposed ~\cite{rusinkiewicz2001efficient,pomerleau2015review,forstner2017efficient}, these methods are still sensitive to pose initialization.
Hand-crafted point cloud feature descriptors, like PFH~\cite{rusu2008aligning}, FPFH~\cite{rusu2009fast}, promise to overcome the initialization issue by establishing correspondences between descriptive features. FGR~\cite{zhou2016fast} achieves global registration by optimizing a correspondence-based objective function using point feature descriptors. Teaser~\cite{Yang20tro-teaser, yang2019polynomial} applies a graph-theoretic framework to decouple the pose estimation problem into sub-problems which allows them to solve for point clouds with a large fraction of noisy correspondences. However, these optimization-based methods focus on building one-to-one correspondences, which can be ambiguous for similar feature descriptors. In contrast, our one-to-many matching is less affected by such ambiguities.


Recently, deep learning methods learn point feature descriptors for correspondence establishment~\cite{deng2018ppfnet,zeng20173dmatch,gojcic2019perfect,wang2019deep,ginzburg2022deep}. Many methods apply PointNet~\cite{qi2017pointnet,qi2017pointnet++} architecture to extract features from raw point clouds~\cite{yaoki2019pointnetlk}. For example, PPFNet~\cite{deng2018ppfnet} learns feature descriptors from the point distribution in a local neighborhood. DCP~\cite{wang2019deep} and PRNet~\cite{wang2019prnet} apply DGCNN for feature learning. In addition to learning feature extraction, RPM-Net~\cite{yew2020-RPMNet} learns to predict optimal annealing parameters to obtain soft correspondences. More work have been developed specifically for local partial-to-partial point cloud registration~\cite{Fu2021RGM,jiang2021sampling,wu2021feature,idam,lee2021deeppro}. Some deep learning methods supervise training solely using ground truth point correspondences, without access to the ground truth transformation matrix~\cite{Fu2021RGM, qin2022geometric, predator, choy2020deep}. In contrast, our proposed method does not have any trainable parameters, thus, does not require a large amount of annotated data.

\textbf{Category-level object pose estimation}
Next to work that tries to match point clouds of the same underlying object, the second line of work proposes to align shapes in a given category to a common, learned coordinate frame. Category-level object pose estimation methods achieve impressive results when trained with supervision~\cite{rempe2020caspr, novotny2019c3dpo, wang2019normalized}, and using only self-supervision ~\cite{spezialetti2020learning, sun2021canonicalcapsules, li2021leveraging, sajnani2022_condor, katzir2022shape}. Canonical Capsule~\cite{sun2021canonicalcapsules} uses an auto-encoder framework to learn pose-invariant capsules to represent object parts, which can be further used to reconstruct the object in the learned canonical frame. However, Canonical Capsules only work with complete point clouds. Equi-pose~\cite{li2021leveraging} leverages a SE(3) equivariant network to estimate the object pose and learns an invariant shape reconstruction module that works on both partial and complete point clouds. While these methods achieve impressive results on synthetic datasets, our experiments show that they do not work well on noisy real-world scans.

\begin{figure}[t]
\centering
\includegraphics[width=0.48\textwidth]{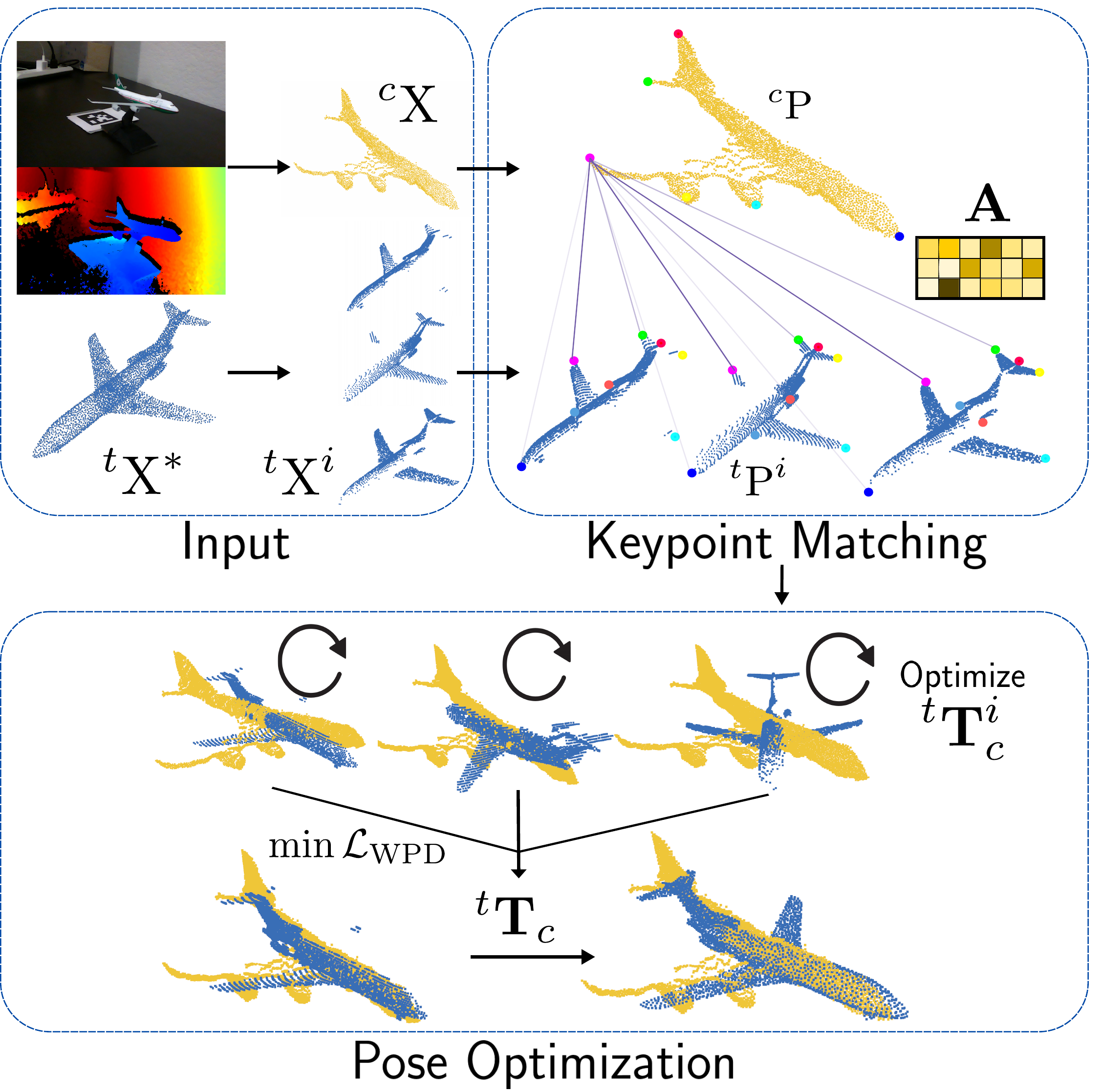}
  \caption{\textbf{Method Overview.} Given a partial point cloud ${}^c\mypcd{X}$ and a complete template point cloud ${}^t\mypcd{X}^*$, we first partition the template into $m=18$ partial point clouds ${}^t\mypcd{X}^{i}$. Our method samples keypoints using farthest point sampling and generates an affinity matrix $\mymat{A}$ that matches each keypoint to points in the partial templates. Using this affinity matrix, we jointly optimize the matching loss and the correspondence matrix to obtain better matches. The optimal pose is the one that minimizes the matching loss between the observed and the partial template point clouds.} \vspace{-0.15in}
\label{fig:method}
\end{figure}

\section{Method} 
\label{method}
We formulate the camera pose estimation problem as matching an observed partial point cloud ${}^c\mypcd{X}$ in the camera frame $\{c\}$ to a complete template point cloud ${}^t\mypcd{X}^*$ from the same category as the observed object, in the template's object frame $\{t\}$. In our notation, the $*$ superscript denotes the complete point cloud. We assume known point normals (which can be computed from the depth images~\cite{nakagawa2015estimating} or through local plane fitting~\cite{hoppe1992surface}), and known ground normal (which can be estimated by upright orientation estimation~\cite{UPRL21,pang2022upright}). In practice, however, we found it sufficient to estimate the ground normal by fitting a plane to the masked background point cloud. Given the observed and complete template point clouds, our objective is to estimate the rigid transformation $^t\mymat{T}_c \in SE(3)$ that transforms the observed point cloud in the camera frame to the template frame.


\subsection{Overview of our approach}
\label{sec:overview}
Given an object's depth image, we generate a partial point cloud ${}^c\mypcd{X}$ by lifting the depth pixels into 3D using the known camera intrinsics. Given this partial point cloud in the camera frame and a template point cloud ${}^t\mypcd{X}^*$, we estimate the rigid transformation $^t\mymat{T}_c$ by following the classical two-step pipeline: correspondence search and pose estimation. Our approach is visualized in Fig.~\ref{fig:method}. We first create a dictionary of $m$ partial template point clouds ${}^t\mypcd{X}^{i} \in \myset{X}$, where $\myset{X} = \{ {}^t\mypcd{X}^{i} | i = 1,\dots,m \}$ is constructed by removing invisible points of ${}^t\mypcd{X}^*$ from $m$ viewpoints. Partitioning is necessary, as feature descriptors on the partial point cloud will not match the descriptors on the complete template due to differences in neighborhood point distributions. Note that this dictionary generating process differs from pose initialization as the partial template point clouds have the same orientation in the template frame; only different parts are missing. Next, we extract keypoints $^c\mypcd{P}$ and $^t\mypcd{P}^i$ from the partial input and the partial template point clouds, respectively. We extract feature descriptors from the point clouds and construct affinity matrices to match the input point cloud ${}^c\mypcd{X}$ to a set of partial point clouds ${}^t\mypcd{X}^{i}$. This matching process generates $m$ pose candidates. We jointly optimize the pose estimate and the correspondence matrix by minimizing the weighted Euclidean distance between keypoints. We select the final pose estimation as the one that minimizes the matching loss between the partial input point cloud and the template.
Next, we describe each component in detail.

\subsection{Affinity matrix from feature matching}\label{sec:feature}
To match a partial point cloud to each partitioned template in the dictionary, we aim to establish an affinity matrix $\mymat{A}$ which indicates the pairwise similarity between the point feature descriptors. First, we extract keypoints $^c\mypcd{P} \in \mathbb{R}^{3 \times n}$, $^t\mypcd{P}^i \in \mathbb{R}^{3 \times 2n}$ from the input and the partial template point cloud respectively using farthest point sampling (FPS)~\cite{qi2017pointnet++}. 

We also require expressive feature descriptors to match the input to the partial templates. In this paper, we provide two options for the point feature descriptor. The first is the Point Feature Histogram (PFH)~\cite{rusu2008aligning}, a commonly used descriptor in the literature, and the second one is a new descriptor that we introduce, named Local Patch Similarity (LPS).

\textbf{Point Feature Histogram (PFH):} 
PFH is a histogram-based, rotation-invariant point feature descriptor that measures the curvature property within the local neighborhood of the keypoint based on the point normals. Once we have extracted PFH descriptors, we can match them by applying Earth Mover's Distance (EMD)~\cite{rubner1998metric} to measure their similarity. Therefore, the affinity score given by the PFH descriptor between two keypoints is:
\begin{equation}
\mymat{A}_{jk} = 1/\big(\text{EMD}(f_{PFH}(^c\mypcd{P}_j),f_{PFH}(^t\mypcd{P}_k))+\epsilon\big)
\end{equation}
where $\epsilon$ is a small value to avoid zero division.

\begin{figure}[t]
\begin{center}
   \includegraphics[width=\linewidth]{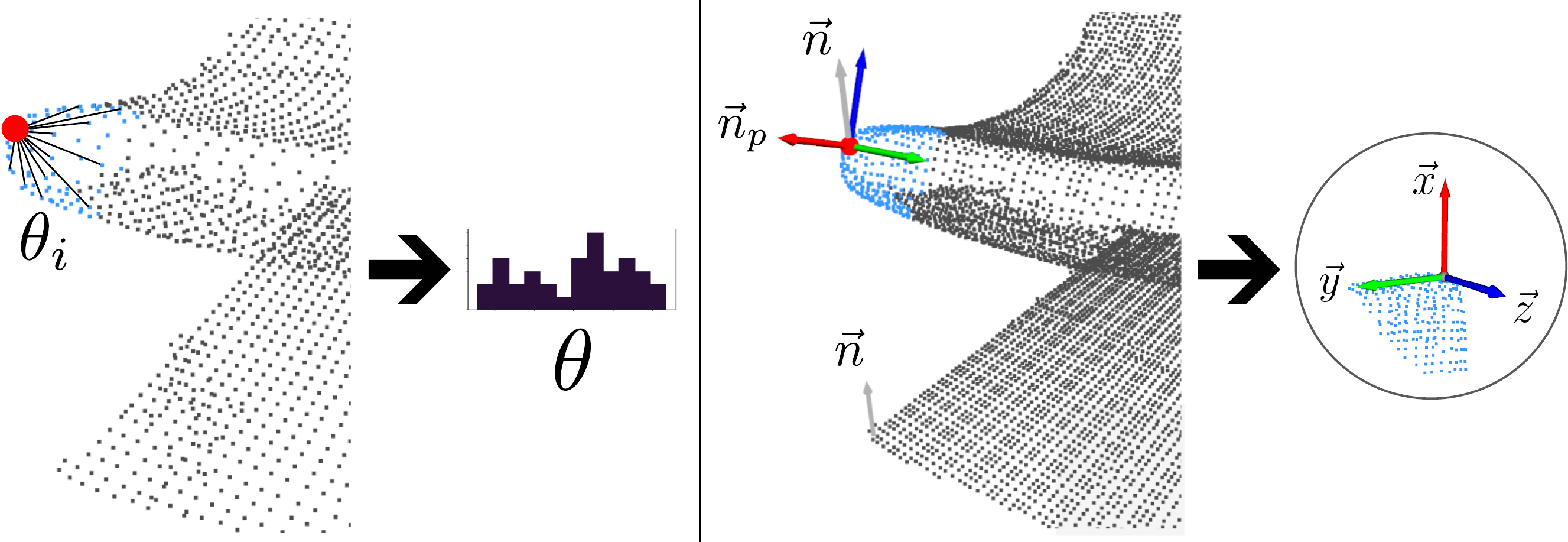}
\end{center}
   \caption{The \textbf{Point Feature Histogram (PFH)} and \textbf{Local Patch Similarity (LPS)} descriptors. PFH (left) uses a histogram of angles to capture the pairwise geometrical properties between all points within a keypoint's neighborhood. LPS (right) constructs a local coordinate frame for the neighborhood of a keypoint, and computes the similarity of two keypoints using the F-score between them.} \vspace{-0.2in}
\label{fig:local_patch}
\vspace{-0.1in}
\end{figure}

\textbf{Local Patch Similarity (LPS):}
For our second descriptor, we introduce the novel LPS descriptor. 
Given a point cloud $\mypcd{X}$, keypoint $p \in \mypcd{X}$, point normal $\myvec{n}_p$, ground normal $\myvec{n}$, and a feature radius $r$, we first define a local reference frame as shown in Fig.~\ref{fig:local_patch}. The $x$-axis of the local reference frame is defined as the point normal $\myvec{x} = \myvec{n}_p$, $y$-axis is defined as $\myvec{y} = \myvec{n} \times \myvec{x}$, and $z$-axis is defined as $\myvec{z} = \myvec{x} \times \myvec{y}$.
In the case when $\myvec{n}$ is aligned with $\myvec{n}_p$, we apply PCA using the points within the neighborhood of $p$ and set $\myvec{x}$ to be the eigenvector with the smallest eigenvalue. Finally, we crop all the points with a distance to $p$ less than the feature radius $r$ and transform them into the local reference frame.

To measure the similarity between two keypoints, we calculate the F-score~\cite{what3d_cvpr19} between two local patches. Therefore, the affinity score given by our feature descriptor is:

\begin{equation}
\mymat{A}_{jk} = \text{F-score}(f_{LPS}(^c\mypcd{P}_j),f_{LPS}(^t\mypcd{P}_k))
\end{equation}

Given these two feature descriptors, we can create our affinity matrix $\mymat{A} \in \mathbb{R}^{n \times 2n}$ by either using one or the other descriptors' affinity scores. Finally, we normalize the affinity matrix $\mymat{A}$ such that each row sums to 1.


\subsection{Initial pose estimation using SVD} \label{sec:method_init}
 The next step is to build pairing points between the two frames for pose initialization. We assign a corresponding point in the template frame for each keypoint in the partial point cloud by weighting all the template keypoints with the associated affinity score. Since the affinity matrix is normalized such that each row sums to $1$, the soft assignment can be achieved by $\mypcd{A}\cdot{}^t\mypcd{P}^\top$. In other words, we \emph{softly} assign a corresponding virtual point in the template frame for each point in $^c\mypcd{P}$ with the noisy affinity matrix $\mymat{A}$. Following the method introduced by~\cite{sorkine2017least, wang2019deep}, the initial pose can be solved as:

\begin{gather}
  \mymat{U} \mymat{\Sigma} \mymat{V}^\top = \text{SVD}({}^c\bar{\mypcd{Q}} \cdot {}^t\bar{\mypcd{Q}}^\top) \\
  ^t\mymat{R}_c = \mymat{V} \mymat{U}^\top \text{and}  \ ^t\myvec{t}_c = -^t\mymat{R}_c{}^c\bar{\mypcd{Q}} + {}^t\bar{\mypcd{Q}}
\end{gather}

where ${}^c\bar{\mypcd{Q}}$ is the translation normalized ${}^c\mypcd{P}$ and ${}^t\bar{\mypcd{Q}}^\top$ is the translation normalized $\mypcd{A}\cdot{}^t\mypcd{P}^\top$. $^t\mymat{R}_c$ and $^t\myvec{t}_c$ are the rotation and translation components of the estimated pose $^t\mymat{T}_c$. The sign of the diagonal matrix $\mymat{\Sigma}$ accounts for the choice of reversing the orientation.


\subsection{Optimize the pose}
\label{sec:optimization}
We first update the estimated pose $^t\mymat{T}_c$ to minimize the following loss function:
\begin{equation}
\mathcal{L}_{\text{WPD}}(\mymat{A}, {}^t\mymat{T}_c) = \sum \mymat{A} \odot  d({}^t\mymat{T}_c \cdot {}^c\mypcd{P},  {}^t\mypcd{P})
\end{equation}
where $d(.,.)$ denotes the pairwise Euclidean distance between two point sets, and $\odot$ denotes the element-wise matrix multiplication. 
This loss function measures the weighted pairwise distance between $^c\mypcd{P}$ and $^t\mypcd{P}$ such that after transforming keypoints $^c\mypcd{P}$ into the template frame, the pair of keypoints $^c\mypcd{P}_j$ and $^t\mypcd{P}_k$ with higher affinity score $\mymat{A}_{jk}$ matters more. Furthermore, we penalize the estimated pose that moves point pairs with high-affinity scores far away from each other.

The second step of this optimizer is to simultaneously update the estimated pose $^t\mymat{T}_c$ and the correspondence matrix $\mymat{C} \in \mathbb{R}^{n \times 2n}$ where each element $\mymat{C}_{jk}$ indicates the level of correspondence between $^c\mypcd{P}_{j}$ and $^t\mypcd{P}_{k}$. We initialize $\mymat{C}$ as the affinity matrix from the previous step. The loss function now becomes:
\begin{equation} \label{eq:6}
\mathcal{L}_{\text{WPD}}(\mymat{C}, {}^t\mymat{T}_c) = \sum \mymat{C} \odot  d({}^t\mymat{T}_c \cdot {}^c\mypcd{P},  {}^t\mypcd{P})
\end{equation}

With this optimization step, we can gradually strengthen the correspondence between the correct matching pair of $^c\mypcd{P}_j$ and $^t\mypcd{P}_k$ by considering the cost of all point pairs under the same rigid transformation. 

\subsection{Dictionary matching}

After matching the partial point cloud ${}^c\mypcd{X}$ to $m$ point clouds in the template dictionary $\myset{X}$, the final pose is selected as the pose that has the minimum loss after the final optimization step. 


  

\newcommand{\relativeposefulltable}{
\begin{table*}[!htbp]
  \centering
  \resizebox{\linewidth}{!}{%
  \begin{tabular}{l c c c c | c c c c | c c c c}
    \toprule
     & \multicolumn{4}{c}{Airplane} &  \multicolumn{4}{c}{Chair} & \multicolumn{4}{c}{Car} \\
    {\textbf{Methods}} & Mean$^{\circ}\downarrow$ & Med.$^{\circ}\downarrow$ & $@10\uparrow$ & $@30\uparrow$ & Mean$^{\circ}\downarrow$ & Med.$^{\circ}\downarrow$ & $@10\uparrow$ & $@30\uparrow$ & Mean$^{\circ}\downarrow$ & Med.$^{\circ}\downarrow$ & $@10\uparrow$ & $@30\uparrow$\\
    
    \hline
Equi-pose         & 3.373            & 1.801            & 0.993         & 0.992         & 13.717           & 3.058            & 0.873         & 0.93          & 93.231           & 58.709           & 0.386         & 0.488         \\
RPM-Net &17.593 &3.233 &0.819 &0.912
        &17.182 &8.455 &0.577 &0.896 
        &49.057 &20.021 &0.318 &0.583\\
\hline
Ours$^\dagger$    & 1.977 & 0.546&0.99&0.994&8.458&1.973&0.876&0.953&64.235&15.044&0.386&0.605 \\
Ours$^*$ & 27.935&5.645&0.760&0.866
&28.264&16.108&0.277&0.756
&92.004&75.108&0.149&0.407\\ \hline
    \bottomrule
  \end{tabular}%
  }
  \caption{ModelNet40 relative pose estimation}
  \label{table:rel_pose_full}
\end{table*}}
\newcommand{\relativepose}{
\begin{table}[!htbp]
  \centering
  \resizebox{\linewidth}{!}{%
  \begin{tabular}{l c c | c c  | c c }
    \toprule
     & \multicolumn{2}{c}{Airplane} &  \multicolumn{2}{c}{Chair} & \multicolumn{2}{c}{Car} \\
    {\textbf{Methods}} & Mean$^{\circ}\downarrow$ & Med.$^{\circ}\downarrow$  & Mean$^{\circ}\downarrow$ & Med.$^{\circ}\downarrow$ & Mean$^{\circ}\downarrow$ & Med.$^{\circ}\downarrow$ \\
    
    \hline
Equi-pose         & 3.373            & 1.801                   & 13.717           & 3.058                  & 93.231           & 58.709           \\
RPM-Net &17.593 &3.233 
        &17.182 &8.455 
        &49.057 &20.021 \\
\hline
Ours$^\dagger$    & 1.977 & 0.546& 8.458 &1.973 &64.235&15.044\\
Ours$^*$ & 27.935&5.645 
&28.264&16.108
&92.004&75.108 \\ \hline
    \bottomrule
  \end{tabular}%
  }
  \caption{Comparison of relative pose estimation between two partial point clouds on ModelNet40. $^\dagger$ Matching with ground truth model. $^*$ Matching with one template for airplane and car, five templates for chair.}
  \label{table:rel_pose_full}
\end{table}}


\newcommand{\realdatatable}{
\begin{table*}[!htbp]
  \centering
  \caption{Pairwise relative pose estimation between two input partial point clouds on real-world scans. Our method applies the PFH descriptor on the airplane and LPS descriptor for both chair and car.}
  \resizebox{\linewidth}{!}{%
  \begin{tabular}{l c c c c | c c c c | c c c c}
    \toprule
     & \multicolumn{4}{c}{Airplane} &  \multicolumn{4}{c}{Chair} & \multicolumn{4}{c}{Car} \\
    {\textbf{Methods}} & Mean$^{\circ}\downarrow$ & Med.$^{\circ}\downarrow$ & $@10\uparrow$ & $@30\uparrow$ & Mean$^{\circ}\downarrow$ & Med.$^{\circ}\downarrow$ & $@10\uparrow$ & $@30\uparrow$ & Mean$^{\circ}\downarrow$ & Med.$^{\circ}\downarrow$ & $@10\uparrow$ & $@30\uparrow$\\
    
    \hline
    ICP         &  70.194 & \underline{7.404} & \underline{0.552} & 0.617       
                & 108.509 & 118.724 & \underline{0.153} & \underline{0.203}
                & 111.746 & 149.675 & 0.087 & 0.240 \\ 
                
FGR             & 104.593 & 114.430 & 0.018 & 0.110       
                & 109.612 & 117.199 & 0.005 & 0.080    
                & 79.185 & 35.478 & 0.231 & 0.482 \\ 

Teaser++        & 75.116 & 64.788 & 0.026 & 0.265       
                & 111.902 & 121.589 & 0.019 &  0.111    
                & \underline{65.832} & \underline{16.470} & 0.256 & \underline{0.607} \\ 
                  
Equi-pose       & \underline{47.446} & \textbf{5.740} & \textbf{0.710} & \underline{0.744}   
                & 112.581 & 120.814 & 0.030 & 0.077     
                & 96.880 & 173.236 & \underline{0.391} & 0.454 \\ 
                    
RPM-net         & 89.448 & 96.913 & 0.246 & 0.361       
                & \underline{99.165} & \underline{101.577} & 0.016 & 0.107 
                & 106.767 & 126.713 & 0.106 & 0.228 \\ 
\hline
Ours            & \textbf{37.032} & 16.155 & 0.261 & \textbf{0.776} &       
                \textbf{28.026} & \textbf{13.424} & \textbf{0.333} & \textbf{0.836} &     
                \textbf{6.643} & \textbf{4.853} & \textbf{0.868} & \textbf{0.958}  \\ 
 \hline
    \bottomrule
  \end{tabular}%
  }
  
  \label{table:real_data}
  \vspace{-0.2in}
\end{table*}}

\newcommand{\compareoptimizerfull}{
\begin{table}[t]
\centering
 \caption{Partial point cloud registration with ground truth complete shape on ModelNet40. Comparison of directly matching the partial to complete point cloud and our proposed dictionary matching with different feature descriptors. }
  \resizebox{\columnwidth}{!}{%
  \begin{tabular}{c l c c | c c | c c}
    \toprule
    \multirow{2}{*}{ \makecell{\textbf{Pipeline} /\\ \textbf{Descriptor}}} & &  \multicolumn{2}{c}{Airplane} &  \multicolumn{2}{c}{Chair} &  \multicolumn{2}{c}{Car}\\
    & {\textbf{Methods}} & Mean$^{\circ}\downarrow$ & Med.$^{\circ}\downarrow$ & Mean$^{\circ}\downarrow$ & Med.$^{\circ}\downarrow$ & Mean$^{\circ}\downarrow$ & Med.$^{\circ}\downarrow$  \\
    
    \hline
\multirow{2}{*}{ \makecell{-}} 

& Equi-pose        & \underline{3.000}  & 1.753  & 13.717  & 3.058 & 97.501 & 174.271  \\
& ICP                & 44.992  & \textbf{0.189}  & 113.896   & 172.167 & 89.762 & 81.487  \\

\hline
\multirow{2}{*}{ \makecell{Direct match \\ FPFH}}
& FGR                & 52.792  & 24.892  & 91.763   & 92.960 & 77.701 & 79.109  \\

& Teaser++         & 46.251  & 20.657  & 90.609  & 94.106 & 65.424 & 42.650  \\
\hline

\multirow{2}{*}{ \makecell{Dictionary \\ FPFH}}

& FGR                 & 4.054  & 1.235  & 13.016  & 2.814 & 71.380 & 26.895\\

& Teaser++          &   {3.354} & 1.066  & \underline{9.496}   & \underline{2.084}  & 63.674   & 22.393\\

\hline
\multirow{3}{*}{ \makecell{Dictionary \\ LPS}}

& FGR                 & 22.538  & 1.181  & 51.911 & 12.665 & 61.975 & 41.239\\

& Teaser++          &{3.699} & {0.485}   &  11.384  & 2.113 & \textbf{30.656} & \textbf{4.886}  \\

& Ours     & \textbf{1.108} & \underline{0.341} & \textbf{4.759}  & \textbf{0.997} & \underline{39.936} & \underline{6.796} \\
 \hline
    \bottomrule
  \end{tabular}%
  } \vspace{-0.2in}
 
  \label{table:optimizer}
\end{table}}
\newcommand{\compareoptimizerfullwithT}{
\begin{table}[!htbp]
\centering
 \caption{Partial point cloud registration with ground truth complete shape on ModelNet40. Comparison of directly matching the partial to complete point cloud and our proposed dictionary matching with different feature descriptors.  }
  \resizebox{\columnwidth}{!}{%
  \begin{tabular}{c l c c | c c | c c}
    \toprule
    \multirow{2}{*}{ \makecell{\textbf{Pipeline} /\\ \textbf{Descriptor}}} & &  \multicolumn{2}{c}{Airplane} &  \multicolumn{2}{c}{Chair} &  \multicolumn{2}{c}{Car}\\
    & {\textbf{Methods}} & Mean$^{\circ}\downarrow$ / Med.$^{\circ}\downarrow$& $\mathcal{L}_T\downarrow$ & Mean$^{\circ}\downarrow$ / Med.$^{\circ}\downarrow$& $\mathcal{L}_T\downarrow$ & Mean$^{\circ}\downarrow$ / Med.$^{\circ}\downarrow$& $\mathcal{L}_T\downarrow$  \\
    
    \hline
\multirow{2}{*}{ \makecell{-}} 
& Equi-pose        & \underline{3.000} / 1.753  & -  & 13.717 / 3.058  & - & 97.501 / 174.271  &  - \\
& ICP                & 44.992 / \textbf{0.189} & - & 113.896 / 172.167 & - & 89.762 / 81.487  & - \\
\hline
\multirow{2}{*}{ \makecell{Direct match \\ FPFH}}
& FGR                & 52.792 / 24.892 & 0.589 & 91.763 / 92.960  & 1.133 & 77.701 / 79.109  & 0.836 \\

& Teaser++         & 46.251 / 20.657  & 0.577 & 90.609 / 94.106  & 1.209 & 65.424 / 42.650  & 0.728 \\
\hline

\multirow{2}{*}{ \makecell{Dictionary \\ FPFH}}

& FGR                 & 4.054 / 1.235  & 0.044 & 13.016 / 2.814  & 0.192 & 71.173 / 27.204 & 0.740\\

& Teaser++          &   {3.354} / 1.066  & \underline{0.038} & \underline{9.496} / \underline{2.084} & \underline{0.143} & 65.663 / 23.564  & 0.682\\

\hline
\multirow{3}{*}{ \makecell{Dictionary \\ LPS}}

& FGR                 & 22.538 / 1.181  & 0.238 & 51.911 / 12.665  & 0.638 & 61.975 / 41.239  & 0.671\\

& Teaser++          &{3.699} / {0.485}  & 0.044 &  11.384 / 2.113  & 0.172 & \textbf{30.656} / \textbf{4.886}  & \textbf{0.369}  \\

& Ours     & \textbf{1.108} / \underline{0.341}  & \textbf{0.016} & \textbf{4.759} / \textbf{0.997} & \textbf{0.083} & \underline{39.936} / \underline{6.796}  &  \underline{0.461}\\
 \hline
    \bottomrule
  \end{tabular}%
  } \vspace{-0.2in}
 
  \label{table:optimizer}
\end{table}}
\newcommand{\matchonetemplateplane}{
\begin{table}[!t]
  \centering
  \caption{Partial point cloud registration with one category-level template on ModelNet40 airplane. } 
  \resizebox{\linewidth}{!}{%
  \begin{tabular}{c l c c c c c }
    \toprule
    \multirow{2}{*}{ \makecell{\textbf{Pipeline} /\\ \textbf{Descriptor}}} & & \multicolumn{5}{c}{Airplane} \\
    & {\textbf{Methods}} & Mean$^{\circ}\downarrow$ & Med.$^{\circ}\downarrow$ & $@10\uparrow$ & $@30\uparrow$ & $\mathcal{L}_T\downarrow$ \\
    
    \hline
\multirow{2}{*}{ \makecell{-}}
& Equi-pose & \textbf{3.000} & \textbf{1.753}  &  \textbf{0.977}  & \textbf{0.996}  & - \\
 &   ICP           &95.93&92.3468&0.327&0.413&-\\
 \hline
\multirow{2}{*}{ \makecell{Direct match \\ FPFH}}
& FGR          &106.056&102.37&0.002&0.035&1.051\\

& Teaser++      &46.296&19.499&0.221&0.665&{0.552} \\
\hline
\multirow{2}{*}{ \makecell{Dictionary \\ FPFH}}
& FGR           &29.831&9.034&0.535&0.819&0.355\\

& Teaser++      &27.766&8.112&0.594&0.863&\underline{0.337} \\
\hline
\multirow{2}{*}{ \makecell{Dictionary \\ PFH}}
& \multirow{2}{*}{Ours}   & \multirow{2}{*}{\underline{20.954}} & \multirow{2}{*}{\underline{4.735}} &\multirow{2}{*}{\underline{0.824}} &\multirow{2}{*}{\underline{0.902}} &\multirow{2}{*}{\textbf{0.263}} \\
 \\
 \hline
    \bottomrule
  \end{tabular}%
  }\vspace{-0.2in}
  \label{table:match_1_template_plane}
\end{table}}


\newcommand{\templatechair}{
\begin{table}[b]
  \centering
  \vspace{-0.1in}
  \caption{Partial point cloud registration with different numbers of category-level templates on ModelNet40 chair. $^\dagger$ PFH descriptor with dictionary. $^\ddagger$ FPFH descriptor with dictionary.}
  \resizebox{\columnwidth}{!}{%
  \begin{tabular}{c c c c c c }
    \toprule
      \multirow{2}{*}{ \makecell{\textbf{Number of} \\ \textbf{templates} }}& & \multicolumn{4}{c}{Chair} \\
     &\textbf{Methods} & Mean$^{\circ}\downarrow$ & Med.$^{\circ}\downarrow$ & $@10\uparrow$ & $@30\uparrow$ \\
\hline

- & Equi-pose   &\textbf{13.717} &\textbf{3.058} &\textbf{0.837} &\textbf{0.957} \\
\hline
\multirow{4}{*}{1}
& ICP              & 129.095 & 169.172 & 0.071 & 0.161 \\
& FGR$^\ddagger$                  & 100.696  & 104.995  & 0.049 & 0.14  \\

& Teaser++$^\ddagger$           & {96.019} & {103.134}   & {0.077}  & {0.239}  \\

& Ours$^\dagger$     & {41.289} & {15.556} & {0.33}  & {0.688}  \\
    \hline
\multirow{4}{*}{ 5}
& ICP              & 51.872 & 32.804 & 0.351 & 0.483 \\
& FGR$^\ddagger$            & 54.930  & 35.536  & 0.1  & 0.439  \\

& Teaser++$^\ddagger$      & {25.793} & {16.401}   & {0.290}  & {0.725}  \\

& Ours$^\dagger$     & \underline{19.520} & \underline{10.664} & \underline{0.467}  & \underline{0.854}  \\
 \hline
    \bottomrule
  \end{tabular}%
  }
  \label{table:templatechair}
\end{table}}

\newcommand{\ablationsvd}{
\begin{table}[!htbp]
  \centering
  \resizebox{\columnwidth}{!}{%
  \begin{tabular}{c l c c c }
    \toprule
    \multicolumn{1}{c}{\textbf{Methods}} & Mean$^{\circ}\downarrow$ & Med.$^{\circ}\downarrow$ & $@10\uparrow$\\
    \hline

Ours    ${}^\ddagger$          & \textbf{1.108} & \textbf{0.341} & \textbf{0.995}\\
Ours (w/o optimization)   ${}^\ddagger$     & 13.469  &    6.958 & 0.719\\
\hline

Ours  ${}^\dagger$        & \textbf{20.954} & \textbf{4.735} & \textbf{0.824}\\
Ours (w/o optimization) ${}^\dagger$     &32.907 & 16.101 & 0.197\\

\hline
    \bottomrule
  \end{tabular}%
  }
  \caption{Registering partial point clouds with and without our optimization step on ModelNet40 airplane category. $\ddagger$ Matching with ground truth model with LPS descriptor. $\dagger$ Matching with one template per category with PFH descriptor. }
  \label{table:ablation-svd}
\end{table}}

\newcommand{\RPM}{
\begin{table}[!htbp]
  \centering
  \resizebox{\columnwidth}{!}{%
  \begin{tabular}{c l c c | c c | c c}
    \toprule
     \multirow{2}{*}{\shortstack{\textbf{Testing} \\ \textbf{Range}}} & \multirow{2}{*}{\shortstack{\textbf{Training} \\ \textbf{Data}}}
     &\multicolumn{2}{c}{Airplane} & \multicolumn{2}{c}{Chair} & \multicolumn{2}{c}{Car} \\
     
    & & Mean$^{\circ}\downarrow$ & Med.$^{\circ}\downarrow$ & Mean$^{\circ}\downarrow$ & Med.$^{\circ}\downarrow$ & Mean$^{\circ}\downarrow$ & Med.$^{\circ}\downarrow$ \\
    
    \hline
 \multirow{2}{*}{full} 
& original              & 87.036 & 88.141 & 81.444 & 83.844 & 91.991 & 105.895\\
& depth scans                 & \textbf{17.593}  & \textbf{3.233}  & \textbf{17.182}  & \textbf{8.455}  & \textbf{49.057} & \textbf{20.021} \\
\hline
\multirow{2}{*}{$<45^\circ$} 
& original              & 5.204  & \textbf{0.849}  &9.806  &\textbf{3.193}  & 13.065 &12.344\\
& depth scans                   &\textbf{4.454}  & 1.839 & \textbf{7.507} &4.577  & \textbf{2.269} & \textbf{5.689}\\
\hline
    \bottomrule
  \end{tabular}%
  }
  \caption{Quantitative comparison of RPM-Net with different training data and testing range on ModelNet40.}
  \label{table:RPM-only}
\end{table}}
\newcommand{\realRPM}{
\begin{table}[!htbp]
  \centering
  \resizebox{\columnwidth}{!}{%
  \begin{tabular}{c l c c | c c | c c}
    \toprule
     \multirow{2}{*}{\shortstack{\textbf{Testing} \\ \textbf{Range}}} & &\multicolumn{2}{c}{Airplane} & \multicolumn{2}{c}{Chair} & \multicolumn{2}{c}{Car} \\
    &\textbf{Methods} & Mean$^{\circ}\downarrow$ & Med.$^{\circ}\downarrow$ & Mean$^{\circ}\downarrow$ & Med.$^{\circ}\downarrow$ & Mean$^{\circ}\downarrow$ & Med.$^{\circ}\downarrow$ \\
    \hline
\multirow{2}{*}{full} &RPM-Net   & 80.91 & 77.122 & 37.299 & 28.535 & 83.907 & 50.007\\
&Ours      & \textbf{48.678}  & \textbf{12.417}  & \textbf{20.510}  & \textbf{17.225}  & \textbf{18.266} & \textbf{12.442} \\
    \hline
\multirow{2}{*}{$<45^\circ$} &RPM-Net   & 57.374 & 26.895 & \textbf{16.429} & 12.960& 14.845& 10.892\\
&Ours      & \textbf{48.848}  & \textbf{11.819}  & 16.809  & \textbf{8.172}  & \textbf{8.832} & \textbf{8.355} \\
\hline
    \bottomrule
  \end{tabular}%
  }
  \caption{Registering real-world data pairs with relative rotation of different range.}
  \label{table:Real_data_small}
\end{table}}


\newcommand{\TYOL}{
\begin{table}[!b]

\centering
\vspace{-0.1in}
 \caption{Camera pose estimation on the TYO-L dataset. Our method applies PFH descriptor as objects are not fixed on the table in this dataset. }
  \resizebox{\columnwidth}{!}{%
  \begin{tabular}{c c | c | c | c | c | c}
    \toprule
     \hline
    
    \textbf{Methods} & \textbf{Objects} & Bottle & Bowl & Can & Plate & Mug \\
    
    \hline
\multirow{4}{*}{ \makecell{ICP}} & Mean$^{\circ}\downarrow$                                     & 106.087 &88.892 & 22.960 & \underline{25.915}& 124.480\\
& Med.$^{\circ}\downarrow$ & 154.767 &121.756 & 17.249 &\textbf{4.162} & 131.630 \\
&$@10\uparrow$ & 0.0 &\underline{0.125} & \underline{0.469} &\underline{0.781} & \underline{0.005} \\
&$@30\uparrow$ & \underline{0.375} &0.344 & 0.563 &\underline{0.875} &\underline{0.039} \\
\hline
\multirow{4}{*}{ \makecell{FGR}} & Mean$^{\circ}\downarrow$                                     & \textbf{28.541} &57.287 & 43.718 &80.613 & 123.431\\
& Med.$^{\circ}\downarrow$ & 5.896 & 53.424 & 53.608 &61.498 & 130.472\\
&$@10\uparrow$ & \underline{0.75} &0.063 & 0.188 &0.063 & \textbf{0.008}\\
&$@30\uparrow$ & \textbf{0.813} &0.25 & 0.344 &0.188 & 0.031\\
 \hline
\multirow{4}{*}{ \makecell{Teaser++}} & Mean$^{\circ}\downarrow$                                & \underline{35.181} &\underline{50.812} & \underline{14.127} &73.672 & \textbf{116.964}\\
& Med.$^{\circ}\downarrow$ & \textbf{3.202} &\underline{43.725} & \underline{10.485} &22.000 & \textbf{124.381}\\
&$@10\uparrow$ & \textbf{0.813} &\textbf{0.156} & 0.438 &0.156 & 0.001\\
&$@30\uparrow$ & \textbf{0.813} &\underline{0.375} & \underline{0.906} &0.563 & \textbf{0.046}\\
 \hline
\multirow{4}{*}{ \makecell{Ours}} & Mean$^{\circ}\downarrow$                & 36.556 & \textbf{36.542} & \textbf{9.506} & \textbf{15.630} & \underline{121.215}\\
& Med.$^{\circ}\downarrow$ & \underline{5.871} & \textbf{31.999} & \textbf{8.398} & \underline{4.797} & \underline{129.402}\\
&$@10\uparrow$ & \underline{0.75} & \textbf{0.156} & \textbf{0.625} & \textbf{0.906} & 0.002\\
&$@30\uparrow$ & \textbf{0.813} & \textbf{0.438} & \textbf{1.0} & \textbf{0.938} &  0.025\\
    \hline
    \bottomrule
  \end{tabular}%
  } 
  \vspace{-0.0in}
  \label{table:TYOL}
\end{table}}

\newcommand{\realdatatablesmall}{
\begin{table}[t]
\centering
 \caption{Pairwise relative pose estimation between two input partial point clouds on real-world scans. Our method applies the PFH descriptor on the airplane and LPS descriptor for chair and car based on experimental results.}
  \resizebox{\columnwidth}{!}{%
  \begin{tabular}{c c | c | c | c | c | c |c }
    \toprule
     \hline
    
    & \textbf{Methods} & ICP & FGR & Teaser++ & Equi-pose & RPM-Net & Ours \\
    
    \hline
\multirow{4}{*}{ \makecell{Airplane}} & Mean$^{\circ}\downarrow$                                     &70.194 &104.593 &75.116 &\underline{47.446} &89.448 &\textbf{37.032} \\
& Med.$^{\circ}\downarrow$ &\underline{7.404} &114.430 &64.788 &\textbf{5.740} & 96.913&16.155 \\
&$@10\uparrow$ &\underline{0.552} &0.018 &0.026 &\textbf{0.710} &0.246 &0.261 \\
&$@30\uparrow$ &0.617 &0.110 &0.265 &\underline{0.744} &0.361 &\textbf{0.776} \\
\hline
\multirow{4}{*}{ \makecell{Chair}} & Mean$^{\circ}\downarrow$                                     &108.509 &109.612 &111.902 &112.581 &\underline{99.165} &\textbf{28.026} \\
& Med.$^{\circ}\downarrow$ &118.724 &117.199 &121.589 &120.814 &\underline{101.577} &\textbf{13.424} \\
&$@10\uparrow$ &\underline{0.153} &0.005 &0.019 &0.030 &0.016 &\textbf{0.333} \\
&$@30\uparrow$ &\underline{0.203} &0.080 &0.111 &0.077 &0.107 &\textbf{0.836} \\
 \hline
\multirow{4}{*}{ \makecell{Car}} & Mean$^{\circ}\downarrow$                                &111.746 &79.185 &\underline{65.832} &96.880 &106.767 &\textbf{6.643} \\
& Med.$^{\circ}\downarrow$ &149.675 &35.478 &\underline{16.470} &173.236 &126.713 &\textbf{4.853} \\
&$@10\uparrow$ &0.087 &0.231 &0.256 &\underline{0.391} &0.106 &\textbf{0.868} \\
&$@30\uparrow$ &0.240 &0.482 &\underline{0.607} &0.454 &0.228 &\textbf{0.958} \\
 \hline

    \hline
    \bottomrule
  \end{tabular}%
  } 
  \vspace{-0.2in}
  \label{table:real_data}
\end{table}}


\newcommand{\datastats}{
\begin{table}[!t]
\centering
 \caption{Real-world datasets statistics. We measure the pair-wise relative pose for asymmetric objects and the relative angle between the main axis in the camera frame and the ground truth object frame for symmetric objects.}
  \resizebox{\columnwidth}{!}{%
  \begin{tabular}{c | c | c | c | c | c | c |c | c }
    \toprule
    \hline
    & Airplane & Chair & Car & Bottle & Bowl & Can & Plate & Mug \\
\hline
Mean$^{\circ}$&94.446 &73.954 &93.315 &78.117 &88.146 &59.598 &97.346 & 125.830 \\
Med.$^{\circ}$&94.240 &94.073 &94.434 &58.550 &66.706 &58.327 & 119.976 & 130.648\\
$@10$& 0.024&0.014 &0.021 &0.0 &0.0 & 0.0 &0.0 & 0.013 \\
$@30$& 0.161&0.198 &0.165 &0.0 &0.0 & 0.0 &0.0 & 0.05 \\
\hline
    \bottomrule
  \end{tabular}%
  } 
  \vspace{-0.1in}
  \label{table:data_stats}
\end{table}}

%
\section{Experiments}
\begin{figure}
    \centering
    \includegraphics[width=\linewidth]{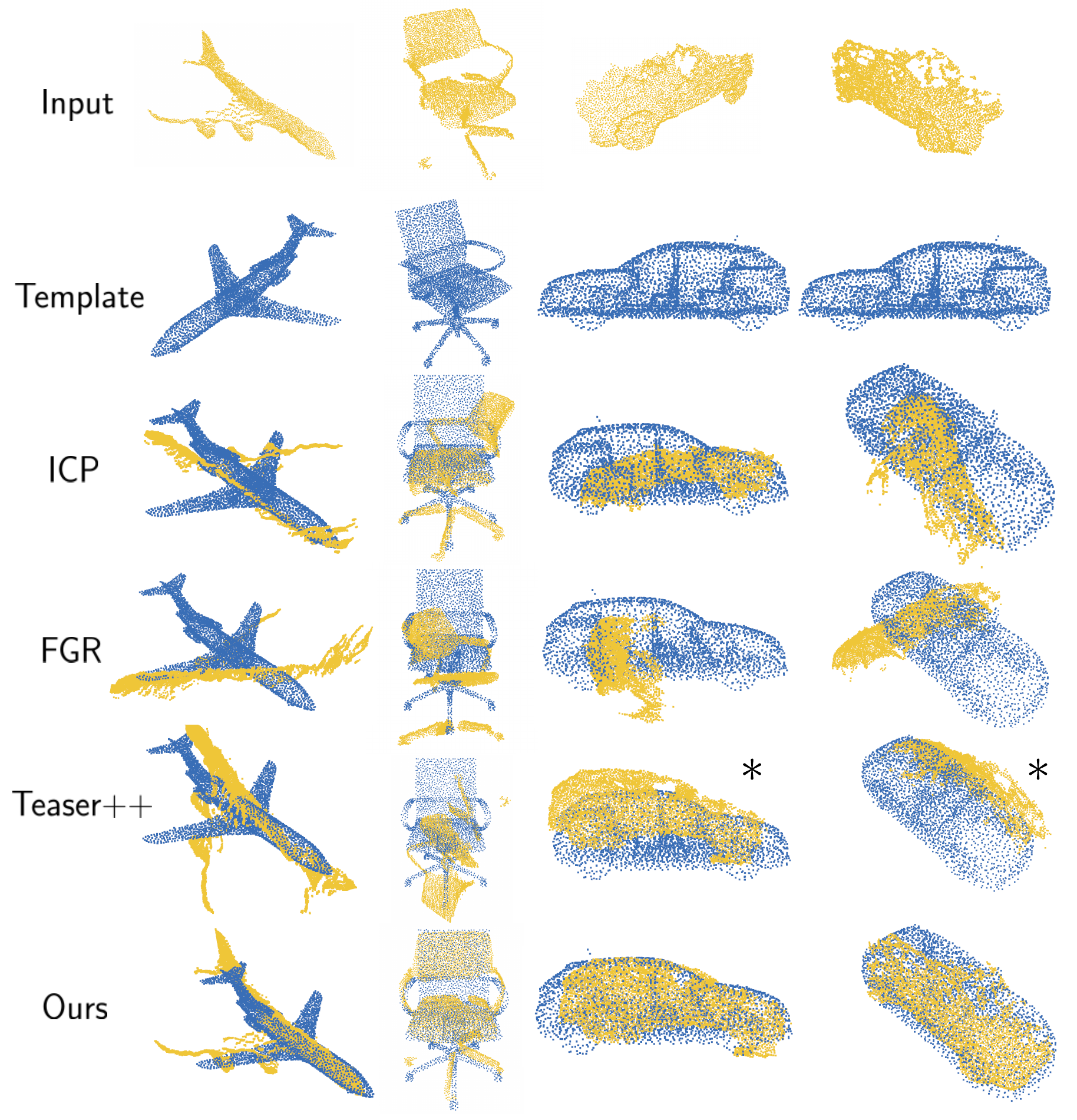}
    \caption{Qualitative results of partial point cloud registration to templates using noisy real-world depth scans. The associated quantitative results can be found in Table~\ref{table:real_data}. Results marked with $^*$ use the affinity matrix computed by our LPS descriptor as Teaser++ did not converge for this car scan with FPFH.}
    \label{fig:real_data_result} \vspace{-0.2in}
\end{figure}

Before showing our result on real-world noisy scans, and our direct applicability to robotic manipulation in Section~\ref{sec:real-world} and~\ref{sec:manipulator}, we design experiments that investigate the limitations of the existing approaches. In Section~\ref{sec:synthetic} we start by registering synthetic partial point clouds to the ground truth complete point cloud, and show in Table~\ref{table:optimizer} how our dictionary matching pipeline improves the baselines compared to directly matching the partial to the complete point cloud. Next, we consider the case when no ground truth complete shapes are available. Therefore, we match the whole synthetic dataset to category-level representative shapes and show that our method outperforms the optimization-based baselines and achieves comparable results with the learning-based methods. In Section~\ref{sec:real-world} and Fig.~\ref{fig:real_data_result}, we present results on challenging noisy real-world scans of both outdoor objects and tabletop objects where no ground truth models are available. We demonstrate that our method outperforms the baselines on a variety of objects and show that optimization-based methods are sometimes more practical in real-world applications when no training process is needed. Finally, we apply our method to real-world robotic grasping and demonstrate that we are able to pick up objects like a mug, cap, can, and headset.

\subsection{Setup}

\textbf{Datasets.}
We conduct experiments on a variety of datasets, including a synthetic dataset ModelNet40~\cite{wu20153d}, a real-world scan benchmark of tabletop objects TYO-L~\cite{hodan2018bop}, and tabletop objects scanned by our manipulator. For ModelNet40~\cite{wu20153d} and TYO-L~\cite{hodan2018bop}, we evaluate all the methods on the official testing split. We also collect a real-world dataset containing three toy airplane models, three outdoor cars, and a chair captured with Intel® RealSense™ cameras. To include more samples, 10 RGBD videos of different chairs from Redwood-3dscan~\cite{Choi2016} are added to this dataset resulting in 17 objects.

\textbf{Implementation details.}
We use the point normal estimation from Open3D~\cite{zhou2018open3d} and the farthest point sampling (FPS) implementation from PyTorch3D~\cite{ravi2020pytorch3d} for keypoint extraction and point cloud down-sampling. We use PyTorch's~\cite{NEURIPS2019_9015} Adam optimizer~\cite{kingma2014adam} to minimize the cost function by updating the estimated pose and correspondence matrix. We set the learning rate to $0.001$ and run 300 steps for each optimization step. We use the continuous 6D representation~\cite{zhou2019continuity} to represent rotations. 

\textbf{Baselines.} 
We compare against optimization-based methods: ICP~\cite{besl1992method}, FGR~\cite{zhou2016fast}, Teaser++~\cite{Yang20tro-teaser}, learning-based partial-to-partial registration method: RPM-Net~\cite{yew2020-RPMNet}, and learning-based canonicalization method: Equi-pose~\cite{li2021leveraging}. For ICP and learning-based methods, we do not report the translation as we shift the input partial point cloud to the zero-mean center. To provide a good initialization for ICP, we divide the full SO(3) into 20 sub-regions and sample one rotation from each sub-region. The learning-based methods are trained on the official training split of ModelNet40 and evaluated on both synthetic and real data of the seen categories.

\textbf{Metrics.}
We perform pose evaluation and report the rotation error in the form of mean, median, $10^{\circ}$ and $30^{\circ}$ accuracy. Given the ground truth pose $\{\mymat{R}_{gt}$, $\myvec{t}_{gt}\}$ and the estimated pose $\{\mymat{R}_{pred}, \myvec{t}_{pred}\}$, the rotation error is defined as:
\begin{equation}
\mathcal{L}_R(\mymat{R}_{gt},\mymat{R}_{pred}) = \arccos{\frac{(\Tr{(\mymat{R}_{gt}^{-1}\mymat{R}_{pred})}-1)}{2}}
\end{equation}
where $\Tr$ is the matrix trace. The translation error is defined as:
\begin{equation}
\mathcal{L}_T(\myvec{t}_{gt},\myvec{t}_{pred}) = ||\myvec{t}_{gt}-\myvec{t}_{pred}||_2
\end{equation}
For all the tables in the following sections, we denote $\mathcal{L}_T$ as the mean translation error for notational simplicity. The best result is in \textbf{bold} and the second best is \underline{underscored}.

\subsection{Point cloud registration on synthetic data} \label{sec:synthetic}

First, we register a partial point cloud to the ground truth model of the same object. As shown in the first four rows of Table~\ref{table:optimizer}, our method outperforms all the other original optimization-based baselines by a large margin. Equi-pose performs well on synthetic data, especially in the airplane and chair categories. Similar to the observation from the original paper~\cite{li2021leveraging}, it predicts many 180-degree-flips when categories are more symmetric, such as in the car category. The tuned ICP achieves a lower median error than ours on the airplane category while having a larger mean error. As expected, ICP achieves a good matching result when the initialization is close to the target pose. Otherwise, it converges to local minima, which leads to large errors. We apply the LPS feature descriptor to our method for this task, while the FGR and Teaser++ use the FPFH descriptor for building correspondences between the partial input and the complete point cloud. 

\compareoptimizerfull

\textbf{Difficulties of directly matching partial to complete point clouds.}
In Section~\ref{sec:overview} we argue that the difference in point distribution between the partial and complete point cloud affects the descriptors, motivating our proposed template partitioning pipeline. 
We justify the need for our pipeline by applying it to FGR and Teaser++ with the original FPFH descriptor. We estimate the transformations between the partial input and all partitioned templates instead of directly registering to the complete shape. Then we select the estimated pose that minimizes the one-directional Chamfer distance from the input to the complete template. As shown in row 5 and 6 of Table~\ref{table:optimizer}, adopting our pipeline significantly improves the performance of FGR and Teaser++.

We further replace the FPFH feature descriptor with our LPS descriptor such that the same affinity matrix is initialized for all three methods. Since the two baselines take only one-to-one correspondences, we assign the correspondences with point pairs of the highest score in the affinity matrix. As shown in Table~\ref{table:optimizer}, our LPS descriptor achieves comparable results with the FPFH descriptor, even with sparse correspondences. Although the point pairs with the highest affinity score given by our LPS descriptor are more likely to be the correct matches, these baselines do not consider multiple hypotheses, which leads to wrong pose estimates if the initial descriptor matching is inaccurate.  

\textbf{Category-level template matching.}
Next, we consider a more realistic scenario where no instance-level ground truth point cloud is available. Therefore, we register the partial input to a template shape chosen from the same category. In Table~\ref{table:match_1_template_plane}, we report the results of matching the entire airplane category to only one template shape. Compared to matching with the ground truth point clouds, we can see a performance drop for all the optimization-based methods. However, we achieve comparable results with the learning-based method Equi-pose. Moreover, we again demonstrate how our matching pipeline improves the baselines in this challenging task. Therefore, we adopt our matching pipeline to FGR and Teaser++ for the following experiments. 
\matchonetemplateplane

We further investigate how multiple templates can help improve the performance of matching. We focus on the chair category, which has the largest shape variety. To select templates, we apply K-means clustering on the training set using the F-score metric and select five representative chair models as our templates. Then the entire test set is registered to these five templates. We report the minimum matching error across templates for all methods. As shown in Table~\ref{table:templatechair}, we can see how all optimization-based methods benefit from adequately selected templates. Moreover, our method outperforms the other optimization-based baselines and achieves comparable results with the learning-based method in this challenging setup.
\templatechair 

\subsection{Camera pose estimation on real-world noisy scans} \label{sec:real-world}
For real-world applications where the object frame is usually unknown, we are interested in estimating the relative camera pose between two depth scans by matching each view to the synthetic category-level template shape. We present the result on our noisy scans of airplane models, chairs, and cars in Table~\ref{table:real_data} and Fig.~\ref{fig:real_data_result}. Another learning-based baseline: RPM-Net, which directly learns partial-to-partial point cloud registration, is also included in this experiment. Since the selected car template is similar to the observed cars, we observe that adding an additional one-directional Chamfer distance to~(\ref{eq:6}) aids the performance. Despite the noise in the depth images, the tuned ICP achieves a lower median rotation error in the airplane category but has a higher mean error. Again, this shows that ICP relies heavily on initialization and is prone to local minima. Equi-pose achieves the lowest median in the airplane category but performs worse in other categories. We also observe that RPM-Net performs better on image pairs with small relative transformations but does not work well on the full dataset. In contrast, our method is more stable for all objects when given a similar template shape to the scan.
\realdatatablesmall

\textbf{Tabletop objects.}
Five categories: bottle, bowl, can, plate, and mug from the TYO-L~\cite{hodan2018bop} dataset are included in this experiment. For each category, we select one object from the Shapenet~\cite{shapenet2015} dataset as the template shape. We report the error in the alignment of the main axis for symmetric objects and relative pose error for asymmetric objects. As shown in Table~\ref{table:TYOL}, our method outperforms the baselines on most objects, while all the feature descriptor-based methods perform similarly on the bottle. We observe that the handles of the mugs are sometimes invisible or too noisy such that all the methods perform poorly on the mug category. However, in the next section, we show that we are able to match the body part of the mug when given a cleaner scan. Finally, we present the datasets statistics for all objects in Table~\ref{table:data_stats}.

\TYOL
\datastats
\subsection{Grasping application} \label{sec:manipulator}
\begin{figure}
    \centering
    \includegraphics[trim={0 3.1cm 0 2.8cm},clip ,width=\linewidth]{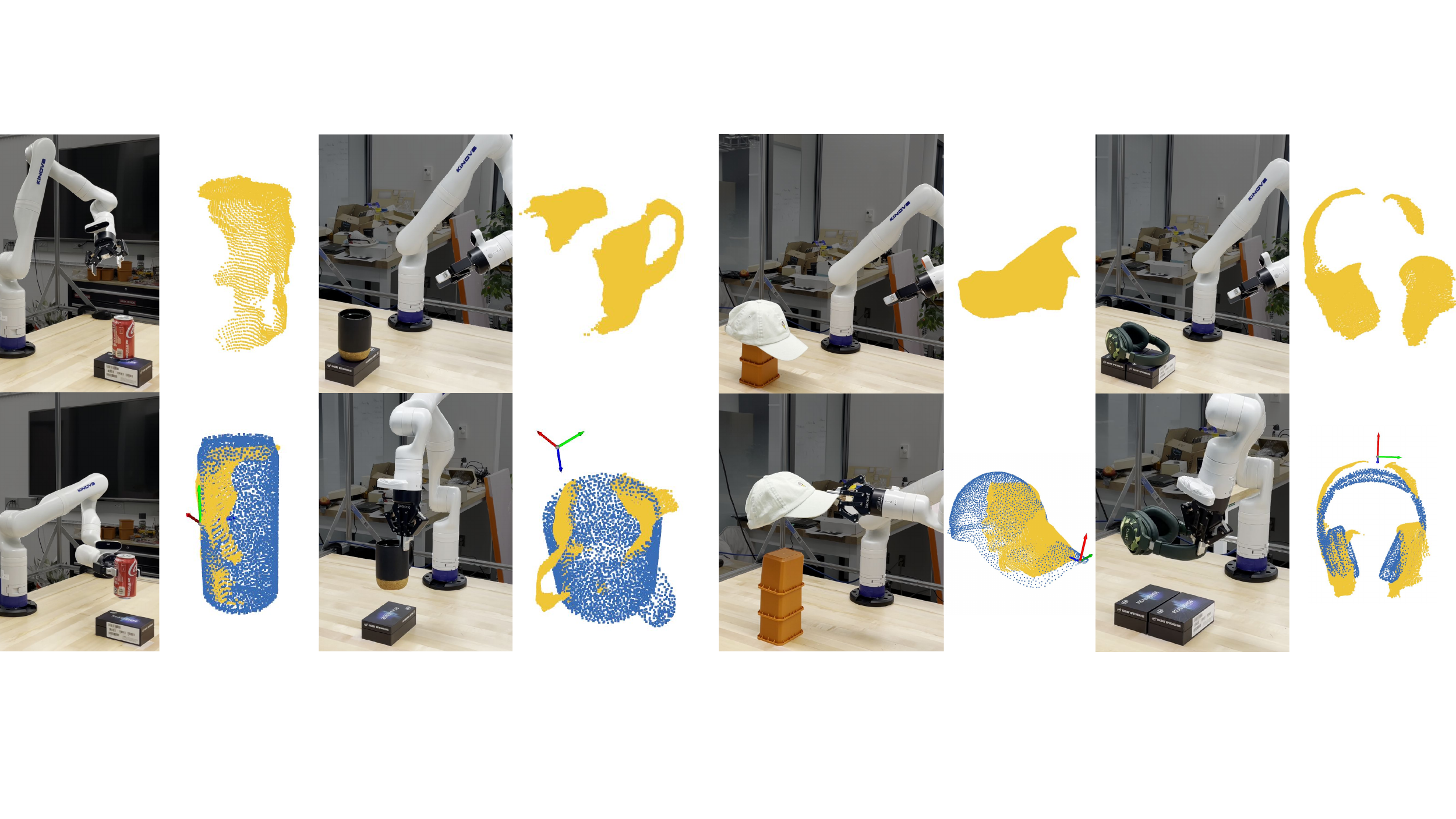}
    \caption{Robotic manipulation experiment. By registering the yellow observed input to the blue template shape, the grasping pose defined in the template frame can be transformed into the camera frame such that the relative gripper pose can be obtained.}
    \label{fig:manipulator_result} \vspace{-0.25in}
\end{figure}

We conduct experiments using a Kinova Gen3 manipulator with a RealSense RGBD camera mounted on the arm. Four tabletop objects: a cap, can, mug, and headset, are included in this experiment. For each object, we select a synthetic shape as the template from the Shapenet~\cite{shapenet2015} dataset and predefine a grasping pose with respect to this template frame. We randomly initialize the manipulator pose such that the camera is roughly pointing at the target object. As shown in the top row of Fig.~\ref{fig:manipulator_result}, given an observed partial point cloud from an unknown pose, we register the observation to the synthetic template shape with our method. In the bottom row of Fig.~\ref{fig:manipulator_result}, we show that after estimating the transformation between the camera frame and the template frame, the relative transformation between the current gripper pose and the grasping pose can be naturally obtained (with the known camera to gripper transformation). Using the estimated relative gripper pose, we show that all four objects can be picked up. Despite the mismatching of the mug's handle, we are still able to pick up the mug along the edge. 

\section{Conclusion, Limitations, and Future Work}
In this paper, we presented a novel optimization method for point cloud registration that considers one-to-many keypoint correspondences together with a new feature descriptor. We showed that our proposed matching pipeline not only outperforms the SoTA but also that components of our method can be used as a drop-in replacement to improve existing methods. Furthermore, we showed that our method is less sensitive to noise and generalizes well on synthetic and real-world data compared to existing methods. Robotic manipulation experiments were also conducted to show that our method applies directly to real-world applications.

One limitation of our method is using 
FPS
for keypoint extraction, a source of randomness. 
Although we showed that FPS is sufficient in our case for keypoint selection, there might be other settings where an additional keypoint detection module could help select more informative keypoints. Moreover, our LPS feature descriptor relies on ground normal estimation, which might not always be accurate. The local reference frame construction retains one degree of freedom ambiguity when the ground normal aligns with the point normal. In future work, we plan to combine our approach with a learned feature descriptor to overcome these issues.



\clearpage
\bibliographystyle{unsrt}
\bibliography{ref}

\begin{thebibliography}{10}

\bibitem{zhang2019eye}
Haotian Zhang, Gaoang Wang, Zhichao Lei, and Jenq-Neng Hwang.
\newblock Eye in the sky: Drone-based object tracking and 3d localization.
\newblock In {\em Proceedings of the 27th ACM International Conference on
  Multimedia}, pages 899--907, 2019.

\bibitem{correll2016analysis}
Nikolaus Correll, Kostas~E Bekris, Dmitry Berenson, Oliver Brock, Albert Causo,
  Kris Hauser, Kei Okada, Alberto Rodriguez, Joseph~M Romano, and Peter~R
  Wurman.
\newblock Analysis and observations from the first amazon picking challenge.
\newblock {\em IEEE Transactions on Automation Science and Engineering},
  15(1):172--188, 2016.

\bibitem{zollini2020uav}
Sara Zollini, Maria Alicandro, Donatella Dominici, Raimondo Quaresima, and
  Marco Giallonardo.
\newblock Uav photogrammetry for concrete bridge inspection using object-based
  image analysis (obia).
\newblock {\em Remote Sensing}, 12(19):3180, 2020.

\bibitem{lowe2004distinctive}
David~G Lowe.
\newblock Distinctive image features from scale-invariant keypoints.
\newblock {\em International Journal of Computer Vision}, 60(2):91--110, 2004.

\bibitem{besl1992method}
Paul~J Besl and Neil~D McKay.
\newblock Method for registration of 3-d shapes.
\newblock In {\em Sensor fusion IV: control paradigms and data structures},
  volume 1611, pages 586--606. Spie, 1992.

\bibitem{segal2009generalized}
Aleksandr Segal, Dirk Haehnel, and Sebastian Thrun.
\newblock Generalized-icp.
\newblock In {\em Robotics: science and systems}, volume~2, page 435. Seattle,
  WA, 2009.

\bibitem{rusu2008aligning}
Radu~Bogdan Rusu, Nico Blodow, Zoltan~Csaba Marton, and Michael Beetz.
\newblock Aligning point cloud views using persistent feature histograms.
\newblock In {\em 2008 IEEE/RSJ international conference on intelligent robots
  and systems}, pages 3384--3391. IEEE, 2008.

\bibitem{rusu2009fast}
Radu~Bogdan Rusu, Nico Blodow, and Michael Beetz.
\newblock Fast point feature histograms (fpfh) for 3d registration.
\newblock In {\em 2009 IEEE international conference on robotics and
  automation}, pages 3212--3217. IEEE, 2009.

\bibitem{zhou2016fast}
Qian-Yi Zhou, Jaesik Park, and Vladlen Koltun.
\newblock Fast global registration.
\newblock In {\em European conference on computer vision}, pages 766--782.
  Springer, 2016.

\bibitem{schwarz2015rgb}
Max Schwarz, Hannes Schulz, and Sven Behnke.
\newblock Rgb-d object recognition and pose estimation based on pre-trained
  convolutional neural network features.
\newblock In {\em 2015 IEEE international conference on robotics and automation
  (ICRA)}, pages 1329--1335. IEEE, 2015.

\bibitem{kendall2015posenet}
Alex Kendall, Matthew Grimes, and Roberto Cipolla.
\newblock Posenet: A convolutional network for real-time 6-dof camera
  relocalization.
\newblock In {\em Proceedings of the IEEE international conference on computer
  vision}, pages 2938--2946, 2015.

\bibitem{kendall2017geometric}
Alex Kendall and Roberto Cipolla.
\newblock Geometric loss functions for camera pose regression with deep
  learning.
\newblock In {\em Proceedings of the IEEE conference on computer vision and
  pattern recognition}, pages 5974--5983, 2017.

\bibitem{tremblay2018deep}
Jonathan Tremblay, Thang To, Balakumar Sundaralingam, Yu~Xiang, Dieter Fox, and
  Stan Birchfield.
\newblock Deep object pose estimation for semantic robotic grasping of
  household objects.
\newblock {\em arXiv preprint arXiv:1809.10790}, 2018.

\bibitem{tekin2018real}
Bugra Tekin, Sudipta~N Sinha, and Pascal Fua.
\newblock Real-time seamless single shot 6d object pose prediction.
\newblock In {\em Proceedings of the IEEE Conference on Computer Vision and
  Pattern Recognition}, pages 292--301, 2018.

\bibitem{suwajanakorn2018discovery}
Supasorn Suwajanakorn, Noah Snavely, Jonathan Tompson, and Mohammad Norouzi.
\newblock Discovery of latent 3d keypoints via end-to-end geometric reasoning.
\newblock {\em arXiv preprint arXiv:1807.03146}, 2018.

\bibitem{pavlakos20176}
Georgios Pavlakos, Xiaowei Zhou, Aaron Chan, Konstantinos~G Derpanis, and
  Kostas Daniilidis.
\newblock 6-dof object pose from semantic keypoints.
\newblock In {\em 2017 IEEE international conference on robotics and automation
  (ICRA)}, pages 2011--2018. IEEE, 2017.

\bibitem{yuan2020deepgmr}
Wentao Yuan, Benjamin Eckart, Kihwan Kim, Varun Jampani, Dieter Fox, and Jan
  Kautz.
\newblock Deepgmr: Learning latent gaussian mixture models for registration.
\newblock In {\em European conference on computer vision}, pages 733--750.
  Springer, 2020.

\bibitem{ginzburg2022deep}
Dvir Ginzburg and Dan Raviv.
\newblock Deep confidence guided distance for 3d partial shape registration.
\newblock {\em arXiv preprint arXiv:2201.11379}, 2022.

\bibitem{wang2019deep}
Yue Wang and Justin~M Solomon.
\newblock Deep closest point: Learning representations for point cloud
  registration.
\newblock In {\em Proceedings of the IEEE/CVF international conference on
  computer vision}, pages 3523--3532, 2019.

\bibitem{yew2020-RPMNet}
Zi~Jian Yew and Gim~Hee Lee.
\newblock Rpm-net: Robust point matching using learned features.
\newblock In {\em Conference on Computer Vision and Pattern Recognition
  (CVPR)}, 2020.

\bibitem{wu2021feature}
Bingli Wu, Jie Ma, Gaojie Chen, and Pei An.
\newblock Feature interactive representation for point cloud registration.
\newblock In {\em Proceedings of the IEEE/CVF International Conference on
  Computer Vision}, pages 5530--5539, 2021.

\bibitem{lee2021deeppro}
Donghoon Lee, Onur~C Hamsici, Steven Feng, Prachee Sharma, and Thorsten
  Gernoth.
\newblock Deeppro: Deep partial point cloud registration of objects.
\newblock In {\em Proceedings of the IEEE/CVF International Conference on
  Computer Vision}, pages 5683--5692, 2021.

\bibitem{Fu2021RGM}
Kexue Fu, Shaolei Liu, Xiaoyuan Luo, and Manning Wang.
\newblock Robust point cloud registration framework based on deep graph
  matching.
\newblock {\em Internaltional Conference on Computer Vision and Pattern
  Recogintion (CVPR)}, 2021.

\bibitem{yaoki2019pointnetlk}
Yasuhiro Aoki, Hunter Goforth, Rangaprasad Arun~Srivatsan, and Simon Lucey.
\newblock Pointnetlk: Robust \& efficient point cloud registration using
  pointnet.
\newblock In {\em The IEEE Conference on Computer Vision and Pattern
  Recognition (CVPR)}, June 2019.

\bibitem{rusinkiewicz2001efficient}
Szymon Rusinkiewicz and Marc Levoy.
\newblock Efficient variants of the icp algorithm.
\newblock In {\em Proceedings third international conference on 3-D digital
  imaging and modeling}, pages 145--152. IEEE, 2001.

\bibitem{pomerleau2015review}
Fran{\c{c}}ois Pomerleau, Francis Colas, Roland Siegwart, et~al.
\newblock A review of point cloud registration algorithms for mobile robotics.
\newblock {\em Foundations and Trends{\textregistered} in Robotics},
  4(1):1--104, 2015.

\bibitem{forstner2017efficient}
Wolfgang Forstner and Kourosh Khoshelham.
\newblock Efficient and accurate registration of point clouds with plane to
  plane correspondences.
\newblock In {\em Proceedings of the IEEE International Conference on Computer
  Vision Workshops}, pages 2165--2173, 2017.

\bibitem{Yang20tro-teaser}
H.~Yang, J.~Shi, and L.~Carlone.
\newblock {TEASER: Fast and Certifiable Point Cloud Registration}.
\newblock {\em {IEEE} Trans. Robotics}, 2020.

\bibitem{yang2019polynomial}
Heng Yang and Luca Carlone.
\newblock A polynomial-time solution for robust registration with extreme
  outlier rates.
\newblock {\em arXiv preprint arXiv:1903.08588}, 2019.

\bibitem{deng2018ppfnet}
Haowen Deng, Tolga Birdal, and Slobodan Ilic.
\newblock Ppfnet: Global context aware local features for robust 3d point
  matching.
\newblock In {\em Proceedings of the IEEE conference on computer vision and
  pattern recognition}, pages 195--205, 2018.

\bibitem{zeng20173dmatch}
Andy Zeng, Shuran Song, Matthias Nie{\ss}ner, Matthew Fisher, Jianxiong Xiao,
  and Thomas Funkhouser.
\newblock 3dmatch: Learning local geometric descriptors from rgb-d
  reconstructions.
\newblock In {\em Proceedings of the IEEE conference on computer vision and
  pattern recognition}, pages 1802--1811, 2017.

\bibitem{gojcic2019perfect}
Zan Gojcic, Caifa Zhou, Jan~D Wegner, and Andreas Wieser.
\newblock The perfect match: 3d point cloud matching with smoothed densities.
\newblock In {\em Proceedings of the IEEE/CVF conference on computer vision and
  pattern recognition}, pages 5545--5554, 2019.

\bibitem{qi2017pointnet}
Charles~R Qi, Hao Su, Kaichun Mo, and Leonidas~J Guibas.
\newblock Pointnet: Deep learning on point sets for 3d classification and
  segmentation.
\newblock In {\em Proceedings of the IEEE conference on computer vision and
  pattern recognition}, pages 652--660, 2017.

\bibitem{qi2017pointnet++}
Charles~Ruizhongtai Qi, Li~Yi, Hao Su, and Leonidas~J Guibas.
\newblock Pointnet++: Deep hierarchical feature learning on point sets in a
  metric space.
\newblock {\em Advances in neural information processing systems}, 30, 2017.

\bibitem{wang2019prnet}
Yue Wang and Justin~M Solomon.
\newblock Prnet: Self-supervised learning for partial-to-partial registration.
\newblock {\em Advances in neural information processing systems}, 32, 2019.

\bibitem{jiang2021sampling}
Haobo Jiang, Yaqi Shen, Jin Xie, Jun Li, Jianjun Qian, and Jian Yang.
\newblock Sampling network guided cross-entropy method for unsupervised point
  cloud registration.
\newblock In {\em Proceedings of the IEEE/CVF International Conference on
  Computer Vision}, pages 6128--6137, 2021.

\bibitem{idam}
Jiahao Li, Changhao Zhang, Ziyao Xu, Hangning Zhou, and Chi Zhang.
\newblock Iterative distance-aware similarity matrix convolution with
  mutual-supervised point elimination for efficient point cloud registration.
\newblock In {\em European Conference on Computer Vision (ECCV)}, 2020.

\bibitem{qin2022geometric}
Zheng Qin, Hao Yu, Changjian Wang, Yulan Guo, Yuxing Peng, and Kai Xu.
\newblock Geometric transformer for fast and robust point cloud registration.
\newblock In {\em Proceedings of the IEEE/CVF Conference on Computer Vision and
  Pattern Recognition (CVPR)}, pages 11143--11152, June 2022.

\bibitem{predator}
Shengyu Huang, Zan Gojcic, Mikhail Usvyatsov, and Konrad~Schindler
  Andreas~Wieser.
\newblock Predator: Registration of 3d point clouds with low overlap.
\newblock In {\em IEEE Conference on Computer Vision and Pattern Recognition,
  CVPR}, 2021.

\bibitem{choy2020deep}
Christopher Choy, Wei Dong, and Vladlen Koltun.
\newblock Deep global registration.
\newblock In {\em Proceedings of the IEEE/CVF conference on computer vision and
  pattern recognition}, pages 2514--2523, 2020.

\bibitem{rempe2020caspr}
Davis Rempe, Tolga Birdal, Yongheng Zhao, Zan Gojcic, Srinath Sridhar, and
  Leonidas~J Guibas.
\newblock Caspr: Learning canonical spatiotemporal point cloud representations.
\newblock {\em Advances in neural information processing systems},
  33:13688--13701, 2020.

\bibitem{novotny2019c3dpo}
David Novotny, Nikhila Ravi, Benjamin Graham, Natalia Neverova, and Andrea
  Vedaldi.
\newblock C3dpo: Canonical 3d pose networks for non-rigid structure from
  motion.
\newblock In {\em Proceedings of the IEEE/CVF International Conference on
  Computer Vision}, pages 7688--7697, 2019.

\bibitem{wang2019normalized}
He~Wang, Srinath Sridhar, Jingwei Huang, Julien Valentin, Shuran Song, and
  Leonidas~J Guibas.
\newblock Normalized object coordinate space for category-level 6d object pose
  and size estimation.
\newblock In {\em Proceedings of the IEEE/CVF Conference on Computer Vision and
  Pattern Recognition}, pages 2642--2651, 2019.

\bibitem{spezialetti2020learning}
Riccardo Spezialetti, Federico Stella, Marlon Marcon, Luciano Silva, Samuele
  Salti, and Luigi Di~Stefano.
\newblock Learning to orient surfaces by self-supervised spherical cnns.
\newblock {\em Advances in Neural Information Processing Systems},
  33:5381--5392, 2020.

\bibitem{sun2021canonicalcapsules}
Weiwei Sun, Andrea Tagliasacchi, Boyang Deng, Sara Sabour, Soroosh Yazdani,
  Geoffrey Hinton, and Kwang~Moo Yi.
\newblock Canonical capsules: Self-supervised capsules in canonical pose.
\newblock In {\em Neural Information Processing Systems}, 2021.

\bibitem{li2021leveraging}
Xiaolong Li, Yijia Weng, Li~Yi, Leonidas Guibas, A~Lynn Abbott, Shuran Song,
  and He~Wang.
\newblock Leveraging se(3) equivariance for self-supervised category-level
  object pose estimation.
\newblock {\em Thirty-Fifth Conference on Neural Information Processing
  Systems}, 2021.

\bibitem{sajnani2022_condor}
Rahul Sajnani, Adrien Poulenard, Jivitesh Jain, Radhika Dua, Leonidas~J.
  Guibas, and Srinath Sridhar.
\newblock Condor: Self-supervised canonicalization of 3d pose for partial
  shapes.
\newblock In {\em The IEEE Conference on Computer Vision and Pattern
  Recognition (CVPR)}, June 2022.

\bibitem{katzir2022shape}
Oren Katzir, Dani Lischinski, and Daniel Cohen-Or.
\newblock Shape-pose disentanglement using se (3)-equivariant vector neurons.
\newblock {\em arXiv preprint arXiv:2204.01159}, 2022.

\bibitem{nakagawa2015estimating}
Yosuke Nakagawa, Hideaki Uchiyama, Hajime Nagahara, and Rin-Ichiro Taniguchi.
\newblock Estimating surface normals with depth image gradients for fast and
  accurate registration.
\newblock In {\em 2015 International Conference on 3D Vision}, pages 640--647.
  IEEE, 2015.

\bibitem{hoppe1992surface}
Hugues Hoppe, Tony DeRose, Tom Duchamp, John McDonald, and Werner Stuetzle.
\newblock Surface reconstruction from unorganized points.
\newblock In {\em Proceedings of the 19th annual conference on computer
  graphics and interactive techniques}, pages 71--78, 1992.

\bibitem{UPRL21}
Luanmin Chen, Juzhan Xu, Chuan Wang, Haibin Huang, Hui Huang, and Ruizhen Hu.
\newblock Learning elastic constitutive material and damping models.
\newblock {\em Computer Graphics Forum (Proceedings of Pacific Graphics)},
  40(7):265--275, 2021.

\bibitem{pang2022upright}
Xufang Pang, Feng Li, Ning Ding, and Xiaopin Zhong.
\newblock Upright-net: Learning upright orientation for 3d point cloud.
\newblock In {\em Proceedings of the IEEE/CVF Conference on Computer Vision and
  Pattern Recognition}, pages 14911--14919, 2022.

\bibitem{rubner1998metric}
Yossi Rubner, Carlo Tomasi, and Leonidas~J Guibas.
\newblock A metric for distributions with applications to image databases.
\newblock In {\em Proceedings of the IEEE/CVF International Conference on
  Computer Vision}, pages 59--66. IEEE, 1998.

\bibitem{what3d_cvpr19}
Maxim Tatarchenko*, Stephan~R. Richter*, René Ranftl, Zhuwen Li, Vladlen
  Koltun, and Thomas Brox.
\newblock What do single-view 3d reconstruction networks learn?
\newblock 2019.

\bibitem{sorkine2017least}
Olga Sorkine-Hornung and Michael Rabinovich.
\newblock Least-squares rigid motion using svd.
\newblock {\em Computing}, 1(1):1--5, 2017.

\bibitem{wu20153d}
Zhirong Wu, Shuran Song, Aditya Khosla, Fisher Yu, Linguang Zhang, Xiaoou Tang,
  and Jianxiong Xiao.
\newblock 3d shapenets: A deep representation for volumetric shapes.
\newblock In {\em Proceedings of the IEEE conference on computer vision and
  pattern recognition}, pages 1912--1920, 2015.

\bibitem{hodan2018bop}
Tomas Hodan, Frank Michel, Eric Brachmann, Wadim Kehl, Anders GlentBuch, Dirk
  Kraft, Bertram Drost, Joel Vidal, Stephan Ihrke, Xenophon Zabulis, et~al.
\newblock Bop: Benchmark for 6d object pose estimation.
\newblock In {\em Proceedings of the European conference on computer vision
  (ECCV)}, pages 19--34, 2018.

\bibitem{Choi2016}
Sungjoon Choi, Qian-Yi Zhou, Stephen Miller, and Vladlen Koltun.
\newblock A large dataset of object scans.
\newblock {\em arXiv:1602.02481}, 2016.

\bibitem{zhou2018open3d}
Qian-Yi Zhou, Jaesik Park, and Vladlen Koltun.
\newblock Open3d: A modern library for 3d data processing.
\newblock {\em arXiv preprint arXiv:1801.09847}, 2018.

\bibitem{ravi2020pytorch3d}
Nikhila Ravi, Jeremy Reizenstein, David Novotny, Taylor Gordon, Wan-Yen Lo,
  Justin Johnson, and Georgia Gkioxari.
\newblock Accelerating 3d deep learning with pytorch3d.
\newblock {\em arXiv:2007.08501}, 2020.

\bibitem{NEURIPS2019_9015}
Adam Paszke, Sam Gross, Francisco Massa, Adam Lerer, James Bradbury, Gregory
  Chanan, Trevor Killeen, Zeming Lin, Natalia Gimelshein, Luca Antiga, Alban
  Desmaison, Andreas Kopf, Edward Yang, Zachary DeVito, Martin Raison, Alykhan
  Tejani, Sasank Chilamkurthy, Benoit Steiner, Lu~Fang, Junjie Bai, and Soumith
  Chintala.
\newblock Pytorch: An imperative style, high-performance deep learning library.
\newblock In H.~Wallach, H.~Larochelle, A.~Beygelzimer, F.~d\textquotesingle
  Alch\'{e}-Buc, E.~Fox, and R.~Garnett, editors, {\em Advances in Neural
  Information Processing Systems 32}, pages 8024--8035. Curran Associates,
  Inc., 2019.

\bibitem{kingma2014adam}
Diederik~P Kingma and Jimmy Ba.
\newblock Adam: A method for stochastic optimization.
\newblock {\em arXiv preprint arXiv:1412.6980}, 2014.

\bibitem{zhou2019continuity}
Yi~Zhou, Connelly Barnes, Jingwan Lu, Jimei Yang, and Hao Li.
\newblock On the continuity of rotation representations in neural networks.
\newblock In {\em Proceedings of the IEEE/CVF Conference on Computer Vision and
  Pattern Recognition}, pages 5745--5753, 2019.

\bibitem{shapenet2015}
Angel~X. Chang, Thomas Funkhouser, Leonidas Guibas, Pat Hanrahan, Qixing Huang,
  Zimo Li, Silvio Savarese, Manolis Savva, Shuran Song, Hao Su, Jianxiong Xiao,
  Li~Yi, and Fisher Yu.
\newblock {ShapeNet: An Information-Rich 3D Model Repository}.
\newblock Technical Report arXiv:1512.03012 [cs.GR], Stanford University ---
  Princeton University --- Toyota Technological Institute at Chicago, 2015.

\end{thebibliography}

\end{document}